\documentclass[conference]{IEEEtran}
\IEEEoverridecommandlockouts
\usepackage{cite}
\usepackage{amsmath,amssymb,amsfonts}
\usepackage{algorithmic}
\usepackage{graphicx}
\usepackage{textcomp}
\usepackage{xcolor}
\def\BibTeX{{\rm B\kern-.05em{\sc i\kern-.025em b}\kern-.08em
    T\kern-.1667em\lower.7ex\hbox{E}\kern-.125emX}}

\newcommand{\donotdisplay}[1]{}
\graphicspath{{Figures/}}
\usepackage{cancel}

\begin{document}

\title{Powered Hawkes-Dirichlet Process: Challenging Textual Clustering using a Flexible Temporal Prior}

\author{\IEEEauthorblockN{Gaël POUX-M\'EDARD}
\IEEEauthorblockA{\textit{ERIC Lab} \\
\textit{Université de Lyon, 69361}\\
Lyon, France \\
0000-0002-0103-8778}
\and
\IEEEauthorblockN{Julien VELCIN}
\IEEEauthorblockA{\textit{ERIC Lab} \\
\textit{Université de Lyon, 69361}\\
Lyon, France \\
0000-0002-2262-045X}
\and
\IEEEauthorblockN{Sabine LOUDCHER}
\IEEEauthorblockA{\textit{ERIC Lab} \\
\textit{Université de Lyon, 69361}\\
Lyon, France \\
0000-0002-0494-0169}
}

\maketitle

\begin{abstract}
The textual content of a document and its publication date are intertwined. For example, the publication of a news article on a topic is influenced by previous publications on similar issues, according to underlying temporal dynamics. However, it can be challenging to retrieve meaningful information when textual information conveys little information or when temporal dynamics are hard to unveil. Furthermore, the textual content of a document is not always linked to its temporal dynamics.
We develop a flexible method to create clusters of textual documents according to both their content and publication time, the Powered Dirichlet-Hawkes process (PDHP). We show PDHP yields significantly better results than state-of-the-art models when temporal information or textual content is weakly informative. The PDHP also alleviates the hypothesis that textual content and temporal dynamics are always perfectly correlated. PDHP allows retrieving textual clusters, temporal clusters, or a mixture of both with high accuracy when they are not. We demonstrate that PDHP generalizes previous work --such as the Dirichlet-Hawkes process (DHP) and Uniform process (UP). Finally, we illustrate the changes induced by PDHP over DHP and UP in a real-world application using Reddit data.
\end{abstract}


\section{Introduction}
Online information is generated at an unprecedented rate. Every minute, 500,000 comments are posted on Facebook, 400 hours of videos are uploaded on Youtube, and 500,000 tweets are published on Twitter. A possible approach to make sense out of this mass of information is to cluster publication events together. Grouping similar publications together help understanding topics of interest or generate summaries of daily news. Many clustering algorithms are based on text similarity, that is, how similar the words of two published documents are \cite{Blei2003LDA, Rathore2018, Bahdanau2015NeuralMT}. Another relevant variable to group information together is the time of publication \cite{Du2012KernelCascade, Blei2006DynamicTopicModel}. For example, two news articles about forest fires might be unrelated if the second article were published years after the first one despite a close lexical similarity. Imagine a news website that publishes a series about history every day at midday. Temporal dynamics would help understand that a publication the next day at midday is likely to be related to previous publications, even if the story (and thus the vocabulary) is different.

Many models that aim at understanding the temporal dynamics of clusters work by selecting a subset of observations according to a temporal sampling function \cite{Amr2008RCRP, Blei2006DynamicTopicModel, Yin2018ShortTextDHP}. However, sampling observations in time implies defining a sampling function that might not correctly model the temporal dynamics at stake. Besides, these works are based on a Dirichlet prior (DP) for clustering. The DP considers counts as a parameter, where a document always counts for 1. It has been argued that such modeling is not fit to account for the arrival of documents in continuous-time settings. In \cite{Du2015DHP}, the authors combine techniques of standard textual clustering with point processes. The idea is to infer the time-sampling function parameters as well as the rest of the model. Explicitly, they derive the Dirichlet-Hawkes process (DHP) prior for documents cluster allocation that takes time as a parameter and yields non-integer counts. It has been argued that this method cannot handle limited cases where text is less informative (e.g., short texts, overlapping vocabularies) \cite{Yin2018ShortTextDHP}. 

Our present work develops the Powered Dirichlet Hawkes process (PDHP) as a mean to handle this case. Besides, we highlight other limiting cases for which DHP fails whereas PDHP yields good results, for instance when temporal information conveys little information (overlapping Hawkes intensities, few observations). We also show there are cases where documents within a textual cluster do not follow the same temporal dynamics, which the DHP is not designed to handle. For instance, an article published by a popular newspaper is unlikely to have the same influence on subsequent similar articles (temporal dynamics) as the same article published by a less popular newspaper. We overcome all these limitations by developing the Powered Dirichlet-Hawkes process, which yields better results than DHP on every dataset considered (up to +0.3 NMI). It also allows us to distinguish textual clusters from \textit{temporal clusters} (documents that follow the same dynamic independently from their content).

Our contributions are listed below:
\begin{itemize}
    \item We highlight and explain the limitations of the DHP prior: it does not handle weakly informative temporal and textual information and it is not designed to consider different dynamics between text and time.
    \item We derive the Powered Dirichlet Hawkes process (PDHP) as a new prior in Bayesian non-parametric for the temporal clustering of a stream textual documents, which is a generalization of the Dirichlet-Hawkes process (DHP) and of the Uniform process (UP).
    \item We show how the PDHP prior performs better than DHP and UP priors through thorough evaluation and comparison on several synthetic datasets and real-world datasets from Reddit.
    \item We show that PDHP prior allows to select the information clusters are based on; we choose to favor their generation more according to documents' textual content or temporal dynamics.
\end{itemize}

\section{Background}
\subsection{How publication times carry valuable information}
Before reviewing existing methods incorporating a temporal dimension into text clustering, we detail how this information is relevant to the task. Recent works on the online spread of textual documents have highlighted several key properties regarding the link between textual content and date of publication.

Firstly, it has been shown that textual documents do not get published independently one from the other. Often, the arrival of a document is conditional on the publication of earlier documents. A straightforward illustration is that a new research paper is built on previous publications and is likely to treat a similar topic; the present article exists because of all the references it cites. A 2012 research paper highlights the critical role played by interactions in the re-publication of a tweet on Twitter \cite{Myers2012CoC}. The authors claim the probability of retweets vary by 71\% on average when considering temporal interactions. More recent works find that although the interaction between publications plays a significant role in later publications, the interaction matrix is often sparse \cite{Poux2021interactions} -- an article on textual clustering is more likely to appear conditional to publications about NLP, whose vocabulary is only a small subset of the scientific literature's one. It highlights the need to cluster words together to retrieve temporal interaction relevant to a textual clustering problem. In this context, a cluster should carry information about the interaction between the documents it contains.

Secondly, a problem that arises is the temporal aspect of interaction. It has been shown that online information interaction decays quickly with time \cite{Cao2019AdsDataset}. Although the rate at which interaction influence decays depends on the dataset, it seems to fade rapidly for most online spreading processes \cite{Haralabopoulos2014LifespanInfo}. To keep the temporal information relevant, clusters must depend on time. For example, two series of news articles about vaccines might not be related (one might not trigger the other) if one was published in 2010 and the other in 2021; they are two different clusters since both obey their own dynamics, although their vocabulary is similar.

\subsection{Temporal clustering of textual documents}
The use of temporal dimension in documents clustering has been studied on many occasions; a notable spike of interest happened in 2006. Many authors tackled the problem of inferring time-dependent clusters from models based on LDA \cite{Blei2006DynamicTopicModel,Wang2006TopicsOverTime,Iwata2009}. However, most of these models are parametric, meaning the number of clusters is fixed at the beginning of the algorithm. Depending on the considered time range and the dataset, the number of clusters needs to be fine-tuned with several independent runs, making them hardly usable for many real-world applications. In all three references cited, the authors mention that a non-parametric version of the model might be derivable.

In 2008, A. Ahmed \textit{\& al} proposed the Recurrent Chinese Restaurant Process (RCRP) as an answer to this problem \cite{Amr2008RCRP}. Instead of considering a fixed-size dataset, this model can handle a stream of documents arriving in chronological order, and the number of clusters is automatically updated. In this model, time is split into episodes to capture the temporal aspect of cluster formation; it considers an integer count of publications within a given time window. A later version of the model from 2010, the Distance-Dependent Chinese Restaurant Process (DD-CRP), tries to alleviate this approximation by replacing fixed-time episodes with a continuous-time sampling function \cite{Blei2010DDCRP}. However, the model still considers integer counts with only their distribution over time changing. Thus, the model is not designed to consider every temporal information in a continuous-time setting.

In 2015, N. Du \textit{\& al} answered this problem by combining the Dirichlet process with the Hawkes process, used to model the appearance of events in a continuous-time setting. The key idea is to replace the counts of a Dirichlet process with the intensity function of the Hawkes process. The resulting Dirichlet-Hawkes process (DHP) is then used as a prior for clustering documents appearing in a continuous-time stream. The inference is realized with a Sequential Monte-Carlo (SCM) algorithm. Following DHP, two articles have been published extending the idea: the Hierarchical Dirichlet Hawkes process (HDHP) \cite{Valera2017HDHP} in 2016 and Indian Buffet Hawkes process in 2018 \cite{Tan2018IBHP}. Another work proposed an EM algorithm for the inference \cite{Xu2017EMDHP} in 2017 (it uses a heuristic method to update the number of clusters and cannot handle a stream of documents). 

A common feature of all the models we mentioned is that they use a non-parametric Dirichlet process (DP) prior or variations built on it, such as DHP and HDHP. Yet, on several occasions, it has been pointed out that there are no specific reasons to use this process in particular and that alternative forms might work better depending on the dataset. In \cite{Welling2006AlterDP}, the author relaxes several conditions associated with DP and shows that alternative priors are an equally valid choice in Bayesian modeling. In \cite{Wallach2010UnifP}, the authors derive the Uniform process (UP) and show that it performs better on a document clustering task. In \cite{Poux2021PDP}, the authors generalize UP and DP within a more general framework, the Powered Dirichlet process (PDP), and show it performs better than DP on several datasets.

Moreover, it has recently been highlighted that DHP does not work well when the textual information within documents conveys little information, that is when the text is short \cite{Yin2018ShortTextDHP} or when vocabularies overlap significantly. To answer this problem, the authors develop an approach based on Dirichlet process mixtures, which is not designed for continuous-time document streams -- the temporal aspect comes from a sampling function as in \cite{Amr2008RCRP,Blei2010DDCRP}. 
There are other limiting cases for DHP, for instance when temporal information is conveys little information (few observations, overlapping temporal intensities) or when documents within textual clusters do not follow the same temporal dynamics. To overcome those limitations, we develop the Powered Dirichlet-Hawkes process in the next section.

\section{Model and algorithm}
\subsection{Dirichlet prior and alternatives}
We briefly recall the definition of a Dirichlet prior. A Dirichlet prior for clustering implements the assumption that the more a cluster is populated, the more chances a new observation belongs to it (``rich-get-richer'' property).
Besides, there is still a chance that a new observation gets assigned to a newly created cluster. It is often expressed using a metaphor, the Chinese Restaurant process (CRP), and it goes as follows: if an $i^{th}$ client arrives in a Chinese restaurant, they will sit at one of the $K$ already occupied tables with a probability proportional to the number of persons already sat at this table. They can also sit alone at a new table $K+1$ with a probability inversely proportional to the total number of clients in the restaurant. When their choice is made, the next client arrives, and the process is repeated.
Let $c$ be the cluster chosen by the $i^{th}$ customer, $\vec{C^-}$ the table assignment of previous customers up to $i-1$, $N_c$ the population of table $c$, $C$ the number of already occupied tables and $\alpha_0 \in \mathbb{R}^+$ the concentration parameter. The process can be written formally as:
\begin{equation}
\label{eq-CRP}
    \text{CRP} (C_i = c \vert \vec{C^-}, \alpha_0) = 
    \begin{cases}
    \frac{N_c}{\alpha_0 + N} \text{ if c = 1, 2, ..., C}\\
    \frac{\alpha_0}{\alpha_0 + N} \text{ if c = C+1}
    \end{cases}
\end{equation}

The Uniform process \cite{Wallach2010UnifP} has been proposed as an alternative to the DP prior. In this context, a new customer entering the restaurant has an identical chance to sit at either of the occupied tables, and a chance to sit at an empty table inversely proportional to the number of occupied tables. Formally:
\begin{equation}
\label{eq-UP}
    \text{U-CRP} (C_i = c \vert \vec{C^-}, \alpha_0) = 
    \begin{cases}
    \frac{1}{\alpha_0 + C} \text{ if c = 1, 2, ..., C}\\
    \frac{\alpha_0}{\alpha_0 + C} \text{ if c = C+1}
    \end{cases}
\end{equation}

Finally, the Powered Dirichlet process \cite{Poux2021PDP} generalizes the two above, stating that the probability for a new client to sit at a new table depends arbitrarily on the number of customers already sat at this table:
\begin{equation}
\label{eq-PCRP}
    \text{P-CRP} (C_i = c \vert r, \vec{C^-}, \alpha_0) = 
    \begin{cases}
    \frac{N_c^r}{\alpha_0 + \sum_{c'} N_{c'}^r} \text{ if c = 1, 2, ..., C}\\
    \frac{\alpha_0}{\alpha_0 + \sum_{c'} N_{c'}^r} \text{ if c = C+1}
    \end{cases}
\end{equation}
where $r \in \mathbb{R}^+$ is an hyper-parameter. Varying $r$ allows to give more or less importance to the ``rich-get-richer'' hypothesis of DP. Note that $P-CRP (r=0, \vec{C^-}, \alpha_0) = U-CRP(\vec{C^-}, \alpha_0)$ and that $P-CRP(r=1, \vec{C^-}, \alpha_0) = CRP(\vec{C^-}, \alpha_0)$. We will use this more general form in the rest of this work and make $r$ vary to compare those priors in the experimental section.

\subsection{Hawkes processes}
A Hawkes process is defined as a self-stimulating temporal point process. It is used to determine the probability of an event happening given the realization of all previous events in a continuous space. Point processes are fully characterized by the intensity function $\lambda(t)$, which is related to the probability $P$ of an event happening between $t$ and $t+\Delta t$ by $\lambda(t) = \lim_{\Delta t \rightarrow 0} \frac{P(t_{events} \in [t;t+\Delta t])}{\Delta t}$. In the case of Hawkes processes, $\lambda(t)$ is defined conditionally on all the events that happened at times lower than $t$. 
In our setup, we define one Hawkes process for each cluster, independent from the others. The intensity of the Hawkes process associated with cluster $c$ is defined as:
\begin{equation}
\label{eq-HawkesClus}
    \lambda_c(t \vert \mathcal{H}_{<t, c}) = \sum_{\mathcal{H}_{<t, c}} \vec{\alpha_c}^T \cdot \vec{\kappa}(t_{i,c})
\end{equation}
where $t_{i,c}$ is the time of the $i^{th}$ observed in cluster $c$, $\mathcal{H}_{<t, c} = \{t_{i,c} \vert t_{i,c}<t\}_{i=1,2,...}$ is the history of events in cluster $c$ up to $t$, $\vec{\alpha_c}$ is a vector of coefficients, $\vec{\kappa}(t)$ is a vector of kernel functions with the same dimension as $\vec{\alpha}$ and $\cdot$ represents the dot product. The kernel functions are set on stone. We will later infer the weights vector $\vec{\alpha}$ to determine which entries of the kernel vector are the most relevant for a given situation. This technique has become standard in Hawkes processes modeling and used in several occasions \cite{Du2012KernelCascade,Yu2017}. Finally, we consider an additional time-independent Hawkes process (that is a Poisson process) of intensity $\lambda(t) = \lambda_0$. This process is used as the Dirichlet-Hawkes equivalent of the concentration parameter $\alpha_0$ in a Dirichlet process (see Eq.~\ref{eq-CRP}). It translates the probability of opening a new cluster as the realization of a Poisson process. In the same way that in DP no observation is assigned to a cluster whose counts is $\alpha_0$ but instead to a new cluster, no observation will be associated with the Poisson process but instead to a new Hawkes process.
\begin{figure*}
    \centering
    \includegraphics[width=\textwidth]{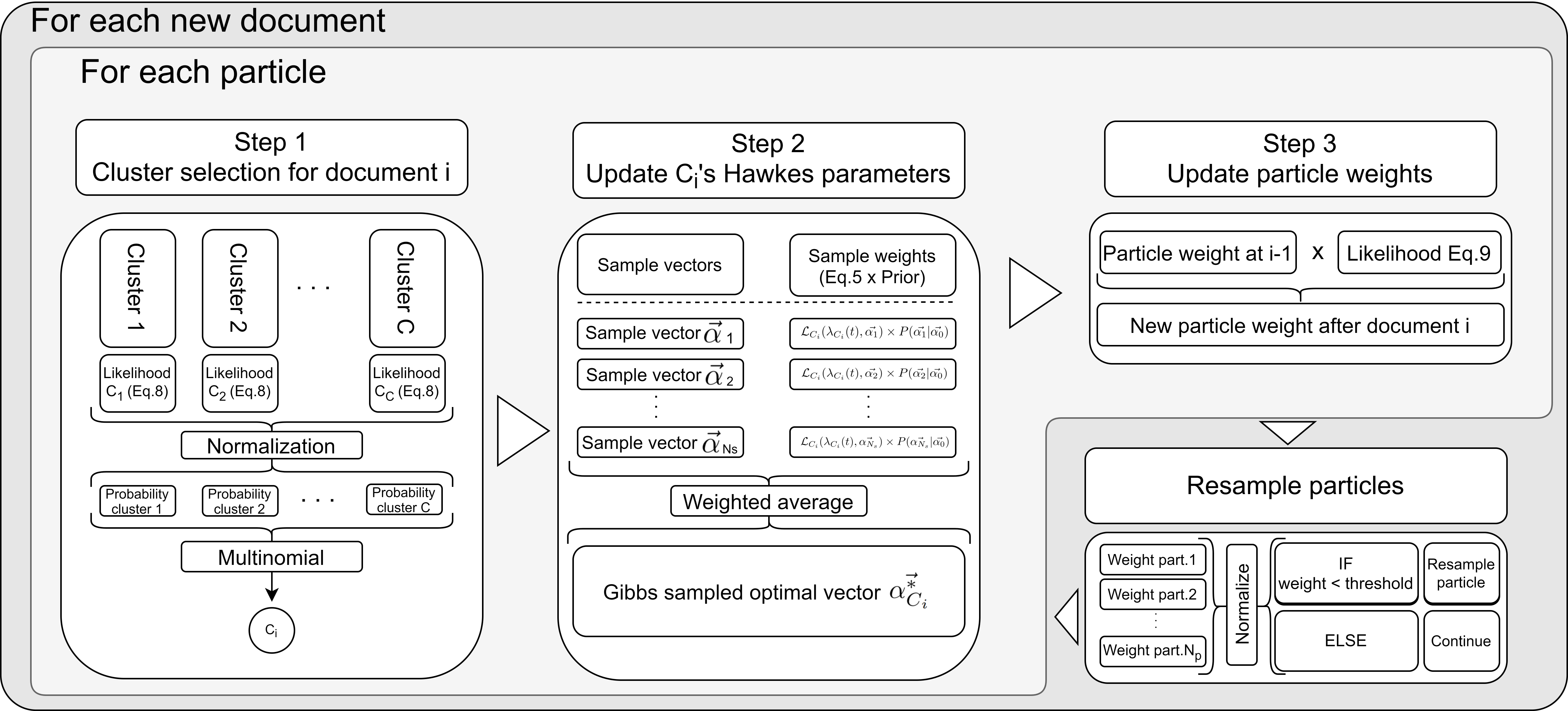}
    \caption{\textbf{Schematic workflow of the SMC algorithm} --- For each new observation from a stream of document, we run steps 1 (sample document's cluster), 2 (update sampled cluster's internal dynamics) and 3 (update particle hypothesis' likeliness) for each particle, and then discard particles containing the less likely hypothesis on cluster allocation.}
    \label{fig-SchemaSMC}
\end{figure*}

Finally, the likelihood of a combination of independent Hawkes processes can be written:
\begin{equation}
    \label{eq-likHawkes}
    \begin{split}
        \mathcal{L}(&\vec{\lambda} \vert \mathcal{H}_{<T, c}) = \mathcal{L}(\lambda_0 \vert \mathcal{H}_{<T, c}) \prod_c \mathcal{L}_c(\lambda_c \vert \mathcal{H}_{<T, c})\\
        &= e^{-\int_0^T \lambda_0 dt} \prod_c e^{-\int_0^T \lambda_c(t)dt}\prod_{t_{i,c}} \lambda_c(t_i \vert \mathcal{H}_{<t_{i,c}, c})\\
        &= e^{-\lambda_0 T + \sum_c \int_0^T \lambda_c(t \vert \mathcal{H}_{<t, c})dt}\prod_{t_{i,c'}, c'=c} \lambda_c(t_{i,c'} \vert \mathcal{H}_{<t_{i,c'}, c})
    \end{split}
\end{equation}
where $T$ is the upper time of the considered observation window, going from $0$ to $T$. Note that $\mathcal{L}(\lambda_0) = e^{-\int_0^T \lambda_0 dt}$ because no event will be assigned to the Poisson process.

\subsection{Powered Dirichlet-Hawkes process}
Following the reasoning in \cite{Du2015DHP}, we substitute the counts $N_k$ of the PDP with the inferred Hawkes intensities in the PDP, resulting in the following form for the Powered Dirichlet-Hawkes prior:
\begin{equation}
\label{eq-likModelTmp}
    P(C_i = c\vert t_i, r, \lambda_0, \mathcal{H}_{<t_i,c}) = 
    \begin{cases}
    \frac{\lambda_c^r(t_i)}{\lambda_0 + \sum_{c'} \lambda_c'^r(t_i)} \text{ if c$\leq$C}\\
    \frac{\lambda_0}{\lambda_0 + \sum_{c'} \lambda_{c'}^r(t_i)} \text{ if c=C+1}
    \end{cases}
\end{equation}
where $t_i$ is the arrival time of document $i$. We reformulated the Dirichlet-Hawkes process in order to allow nonlinear dependence ($r$) on the non-integer counts ($\vec{\lambda}$).

\subsection{Textual modeling}
We choose to model the textual content of documents as the result of a Dirichlet-Multinomial distribution. This model is purposely simple to ease the understanding, but can easily be replaced by a more complex one. A more complete textual modeling is out of the scope of this work, which aims to highlight the efficiency of the PDHP. Here, a document will be associated to a given cluster according to words count in every cluster and words count in the document only. The generative process is as follows:
\begin{equation}
    \theta_i \sim Dir(\theta_0) \ \ \ \ ;\ \ \ \ \omega_{v,i} \sim Mult(\theta_i)
\end{equation}
where $\theta_i$ is the cluster of document $i$, and $\omega_{v,i}$ is the $v^{th}$ word of document $i$. Let $\mathcal{L}_{txt}(\vec{C}_{<i, c} \vert N_{<i,c}, \theta_0)$ be the marginal joint distribution of every document's cluster allocation up to the $i^{th}$ one. The likelihood of the $i^{th}$ document belonging to cluster $c$ can then be expressed as:
\begin{equation}
\label{eq-likModelLg}
    \begin{split}
        \mathcal{L}&(C_i=c \vert N_{<i,c}, n_i, \theta_0) = P(n_i \vert C_i=c, N_{<i,c}, \theta_0)\\
        &= \frac{\mathcal{L}_{txt}(\vec{C}_{<i, c} \vert N_{<i,c}, \theta_0)}{\mathcal{L}_{txt}(\vec{C}_{<i-1, c} \vert N_{<i,c}, \theta_0)} \\
        &= \frac{\frac{\cancel{\Gamma(\theta_0)}}{\Gamma(N_c+n_i+\theta_0)} \prod_v \frac{\Gamma(N_{c,v} + n_{i,v} + \theta_{0,v})}{\cancel{\Gamma(\theta_{0,v})}}}
        {\frac{\cancel{\Gamma(\theta_0)}}{\Gamma(N_c+\theta_0)} \prod_v \frac{\Gamma(N_{c,v} + \theta_{0,v})}{\cancel{\Gamma(\theta_{0,v})}}}\\
        &= \frac{\Gamma(N_c+\theta_0)}{\Gamma(N_c+n_i+\theta_0)} \prod_v \frac{\Gamma(N_{c,v} + n_{i,v} + \theta_{0,v})}{\Gamma(N_{c,v}+\theta_0)}
    \end{split}
\end{equation}
where $N_c$ is the total number of words in cluster $c$ from observations previous to $i$, $n_i$ is the total number of words in document $i$, $N_{c,v}$ the count of word $v$ in cluster $c$, $n_{i,v}$ the count of word $v$ in document $i$ and $\theta_0 = \sum_v \theta_{0,v}$.

\subsection{Posterior distribution}
The resulting posterior distribution of the $i^{th}$ document over clusters is calculated using Bayes theorem. It is proportional to the product of the textual likelihood Eq.\ref{eq-likHawkes} and the temporal Powered Dirichlet-Hawkes prior Eq.\ref{eq-likModelLg}:
\begin{equation}
\label{eq-likModelTot}
\begin{split}
    &P(C_i = c \vert r, n_i, t_i, N_c, \mathcal{H}_{<t, c})\\
    \propto &\underbrace{P(n_i \vert C_i=c, N_{<i,c}, \theta_0)}_{\text{Textual likelihood}} \underbrace{P(C_i = c\vert t_i, r, \lambda_0, \mathcal{H}_{<t_i,c})}_{\text{Temporal prior}} \\
    = &\frac{\Gamma(N_c+\theta_0)}{\Gamma(N_c+n_i+\theta_0)} \prod_v \frac{\Gamma(N_{c,v} + n_{i,v} + \theta_{0,v})}{\Gamma(N_{c,v}+\theta_0)}\\
    &\times \begin{cases}
    \frac{\lambda_c^r(t_i)}{\lambda_0 + \sum_{c'} \lambda_c'^r(t_i)} \text{ if c = 1, ..., C}\\
    \frac{\lambda_0}{\lambda_0 + \sum_{c'} \lambda_{c'}^r(t_i)} \text{ if c = C+1}
    \end{cases}
\end{split}
\end{equation}
We recall that $\lambda_c(t)$ is defined Eq.~\ref{eq-HawkesClus}. The textual likelihood of cluster $C+1$ is computed by setting $N_{C+1,v}=0$.

\subsection{Algorithm and changes induced by PDHP}
We use a similar algorithm to the one in \cite{Du2015DHP}. Briefly, the algorithm is a sequential Monte-Carlo (SMC) that takes one document at a time in their order of arrival. The algorithm starts with a number $N_{part}$ of particles whose weights are $\omega_{p} = \frac{1}{N_{part}}$, each of which will keep track of a hypothesis on documents clusters. After a few iterations, particles that contained unlikely allocation hypotheses are discarded and replaced by more likely ones. The likeliness of a hypothesis is encoded in the weights of each particle $\omega_p$.

For each particle, when a new document arrives, (1) the cluster of the document is sampled according to a Categorical distribution over all clusters, whose weights are determined by Eq.~\ref{eq-likModelTot}. After the cluster of the new document has been sampled, (2) the kernel weights $\vec{\alpha}$ from Eq.~\ref{eq-HawkesClus} are updated using Eq.~\ref{eq-likHawkes}. For efficiency purpose, we infer $\vec{\alpha}$ using Gibbs sampling from a set of $N_s$ pre-computed $\vec{\alpha}$ vectors. We finally (3) update the weights $\omega_p$ of each particle according to the posterior Eq.~\ref{eq-likModelTot} such as $\omega_p^{(n+1)} = \omega_p^{(n)} \times \text{Eq.~\ref{eq-likModelTot}}$. If the weight of a particle falls below a value $\omega_{thres}$, the particle is discarded and replaced by another existing one with sufficient weight. The whole process is illustrated Fig.~\ref{fig-SchemaSMC}. By updating incrementally the likelihood associated with each of the pre-computed $\vec{\alpha}$ sample vectors, the algorithm treats each new observation in constant time $\mathcal{O}(1)$.

\begin{figure}
    \centering
    \includegraphics[width=\columnwidth]{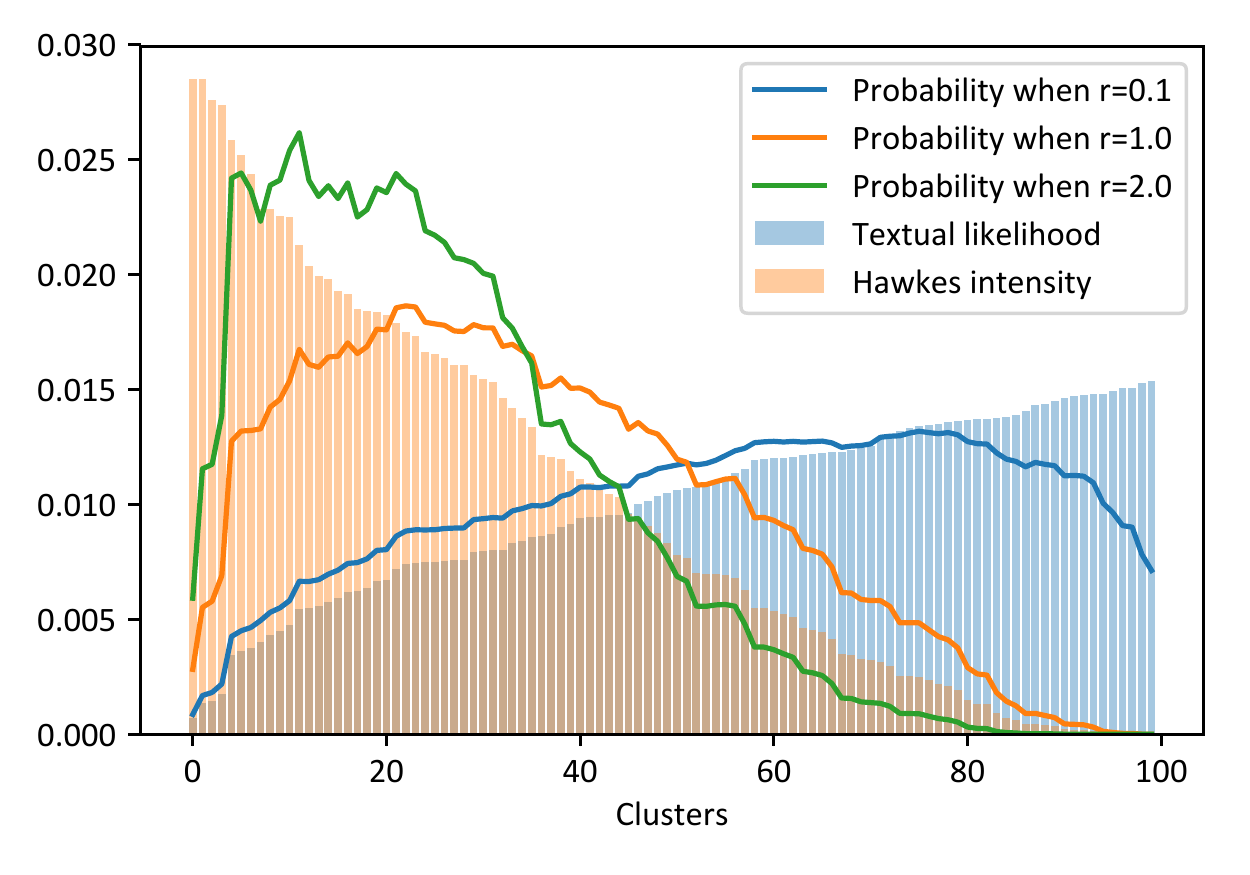}
    \caption{\textbf{Effect of r on cluster selection probabilities} --- The probability for each cluster to get chosen (solid lines) for several values of $r$ and fixed individual textual likelihood (blue bars) and Hawkes intensity (orange bars).}
    \label{fig-ex-varr}
\end{figure}

The task of updating kernels coefficients (2) is the same as in any Hawkes process, and the task of updating particles weights and resampling them (3) is common to any SMC algorithm. The change induced by the PDHP compared to the DHP happens at step (1). First of all, we note that for $r=1$ the PDHP prior is identical to the DHP prior. From \cite{Poux2021PDP}, lowering the value of $r$ reduces the ``rich-get-richer'' aspect of the PDP (``rich-get-less-richer''), whereas increasing it leads to a ``rich-get-more-richer'' effect. These metaphors can be translated as follows in our temporal context: for lower values of $r$, the relative difference between cluster's temporal intensities plays a less important role in cluster selection, whereas higher values of $r$ tend to exacerbate these differences and make the temporal aspect of the greatest consequence on the choice of a cluster. In other words, tuning the value of $r$ allows to give more or less importance to the temporal aspect of the clustering. This is illustrated in Fig.~\ref{fig-ex-varr}, where we plot the probability for various clusters to be chosen (which is directly proportional to the posterior distribution, Eq.~\ref{eq-likModelTot}) according to $r$ when their textual likelihood and Hawkes process intensity is known. Note that for $r=0$, the probability for any cluster to get chosen is directly proportional to its textual likelihood (Dirichlet-Uniform process), whereas when $r$ increases, the probability of getting chosen gets closer to a selection only based on the temporal aspect.

This makes the main interest of the PDHP model. Tuning the parameter $r$ allows one to choose whether inferred clusters are based on textual or temporal considerations. It generalizes several state-of-the-art works, which are special cases of the PDHP for different values of $r$. The DHP \cite{Du2015DHP} is equivalent to PDHP for $r=1$; the UP \cite{Wallach2010UnifP} is equivalent to PDHP when $r=0$.
In the following sections, we show how fine-tuning $r$ systematically yields significantly better results than setting it to $r=0$ or $r=1$ (up to a gain of 0.3 on our experiments' normalized mutual information metric). We also show how varying it allows to recover one kind of clustering or the other (textual or temporal) with high accuracy and see how it affects clustering results on several real-world datasets.

\section{Experiments}
\subsection{Synthetic data generation}
We simulate a case where only two clusters are considered. Each cluster has its own vocabulary distribution over 1~000 words and its own kernel weights $\vec{\alpha}$, with Gaussian Hawkes kernel functions $\vec{\kappa (t)}$ of parameters $(\mu, \sigma)$=(3, 0.5), (7, 0.5) and (11, 0.5) (see Eq.~\ref{eq-HawkesClus}). Finally, we set $\lambda_0=0.05$.
We first simulate one independent Hawkes process per cluster using the Tick Python library \cite{Bacry2017Tick}. The processes are stopped at time $t=1500$, which makes a rough average of 7~000 events per run. Then we associate each simulated observation with a sample of 20 words drawn from the corresponding cluster's word distribution. Inference has been performed using a 8 core processor (i7-7700HQ) with 8GB of RAM on a laptop, which underlines how scalable the algorithm is. As stated before, the algorithm treats each new document in constant time $\mathcal{O}(1)$, which ranged from 0.05s on synthetic data to maximum 1s on real-world data. Note that this number is directly proportional to the number of active inferred clusters, and thus depends strongly on the dataset.

\begin{figure}
    \centering
    \includegraphics[width=\columnwidth]{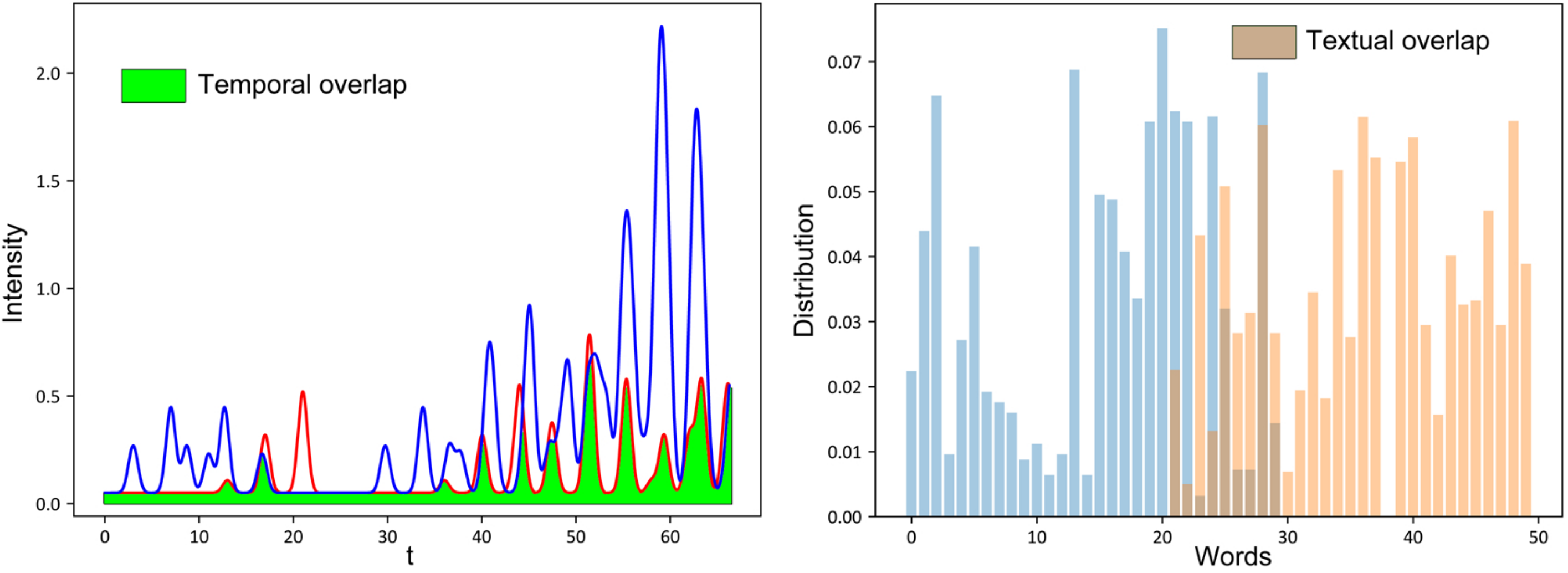}
    \caption{\textbf{Overlaps} --- (Left) Temporal overlap is defined as the ratio between the area common to two Hawkes intensities and the total area under the intensity functions. (Right) Textual overlap is defined as the proportion of vocabulary that is common to two clusters, weighted by the probability of words within their respective cluster.}
    \label{fig-illustration-overlaps}
\end{figure}

We generate ten such datasets for every considered value of vocabulary overlap and Hawkes intensities overlap, which leave us with $\sim$200 datasets. Overlap is defined as the common area of two distributions, normalized by the total area under the distributions. For example, if the vocabulary of one cluster ranges from words "1" to "100" with uniform distribution, and the vocabulary of another cluster from words "50" to "150" with uniform distribution, the overlap equals 50\%. We define the overlap of Hawkes process intensity in the same way. If the triggering Hawkes kernel of one cluster is a Gaussian function with $(\mu, \sigma)=(3, 1)$ and one associated observation at $t=0$, and the triggering kernel of the other is also a Gaussian function but with $(\mu, \sigma)=(5, 1)$ also with an associated observation at $t=0$, the overlap equals 32\% (see Fig.~\ref{fig-illustration-overlaps}). When computing the Hawkes intensity overlap, every observation within a cluster and its associated timestamp are considered. The definition of overlaps is illustrated in Fig.~\ref{fig-illustration-overlaps}. To enforce a given vocabulary overlap (Fig.~\ref{fig-illustration-overlaps}-right), we shift the word distributions of the clusters from which events' vocabulary is sampled. To enforce a given Hawkes intensities overlap (Fig.~\ref{fig-illustration-overlaps}-left), we shift the event times of every event in one of the clusters until we get the correct overlap ($\pm 5\%$).

Note that we consider ten different datasets instead of considering ten runs per dataset for two reasons. Firstly, the generation of Hawkes processes is highly stochastic, so a model might perform significantly better on a single dataset only by chance. Secondly, given the way the SMC algorithm works, the standard deviation between runs is small: at each iteration, $N_{part}$ clustering hypotheses are tested, which is equivalent to running $N_{part}$ times a single clustering algorithm. We heuristically set $N_{part}=8$, as we observe no significant improvement using more particles.

The other parameters we use for clustering synthetic data are: $\alpha_0=0.1$, $\theta_0=1$, $\vec{\kappa(t)} = [\mathcal{G}(t; 3, 0.5), \mathcal{G}(t; 7, 0.5), \mathcal{G}(t; 11, 0.5)]$ with $\mathcal{G}(t; \mu, \sigma)$ the Gaussian function, $N_{samples}=2.000$ and $\omega_{thres}=\frac{1}{2N_{part}}$.\footnote{All codes and implementations are available at https://github.com/GaelPouxMedard/PDHP}
\begin{figure}
    \centering
    \includegraphics[width=\columnwidth]{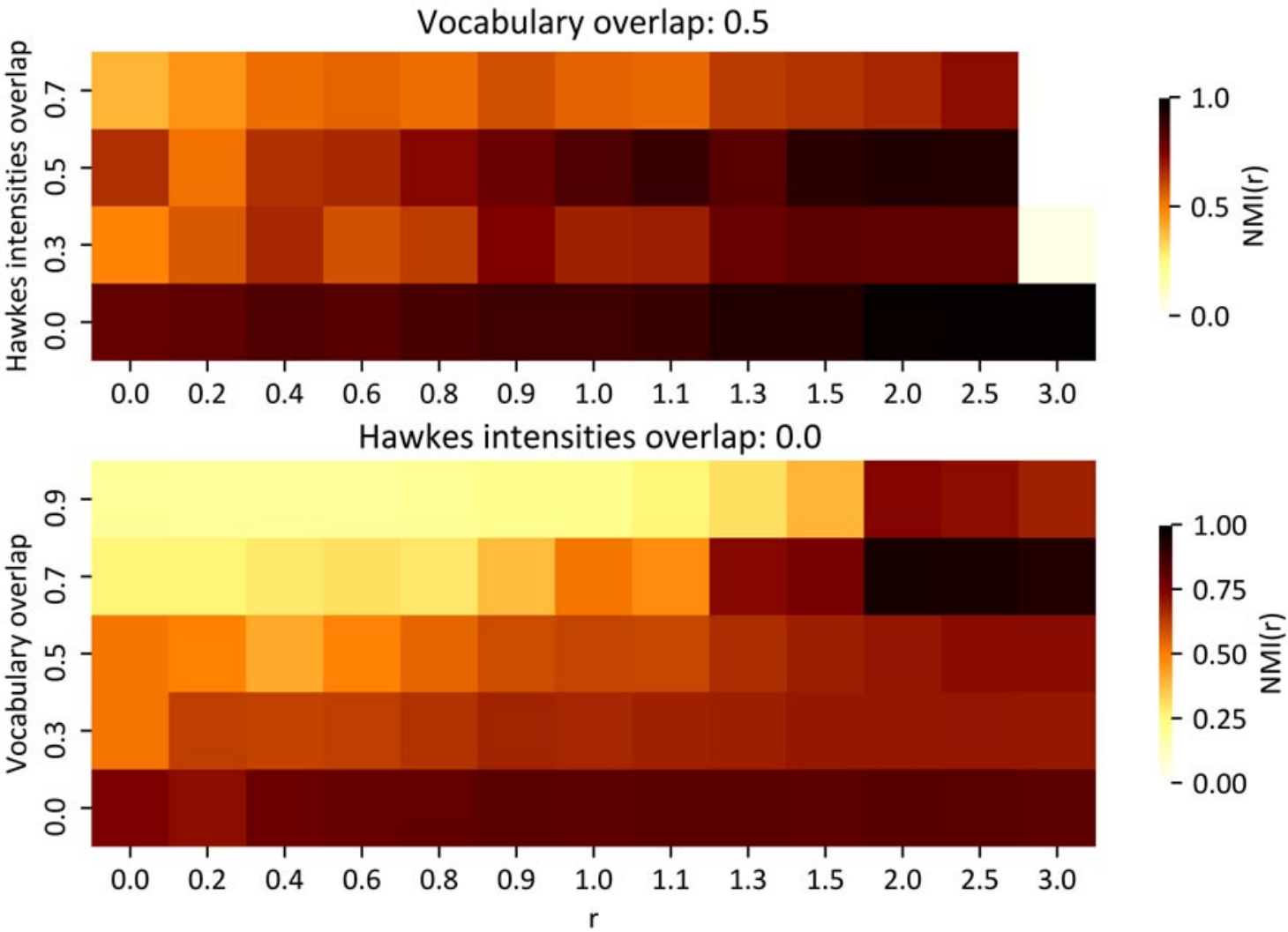}
    \caption{\textbf{PDHP yields good NMI values} --- Normalized mutual information (NMI) for various values of $r$, intensities overlap and vocabulary overlap, for one dataset per combination. The results for $r=0$ are the output of the Uniform process, the results for $r=1$ are the output of the DHP \cite{Du2015DHP}, and the other values of $r$ correspond to other special cases of PDHP. The darker the better. Overall, PDHP yields good values of NMI.}
    \label{fig-res-NMI}
\end{figure}

We are interested in varying both vocabulary and intensities overlap to exhibit the limits of DHP and how PDHP overcomes them. Note that in the synthetic data experiments in \cite{Du2015DHP} (Figs.3a and 3b), the intensities overlap is almost null, which makes the task easier for the Hawkes part of the algorithm. The primary metric we use throughout the experimental section is the normalized mutual information (NMI). During the experiments, we also considered the Adjusted normalized rand index and the V-measure, which are well adapted to evaluate clustering results when the number of inferred clusters is different from the true number of clusters. The observed trends in results from these other metrics are identical to the ones observed for NMI. Therefore we choose to report only the results of the latter for clarity. These additional measurements are provided in the linked repository along with the code and datasets.

\begin{figure}
    \centering
    \includegraphics[width=\columnwidth]{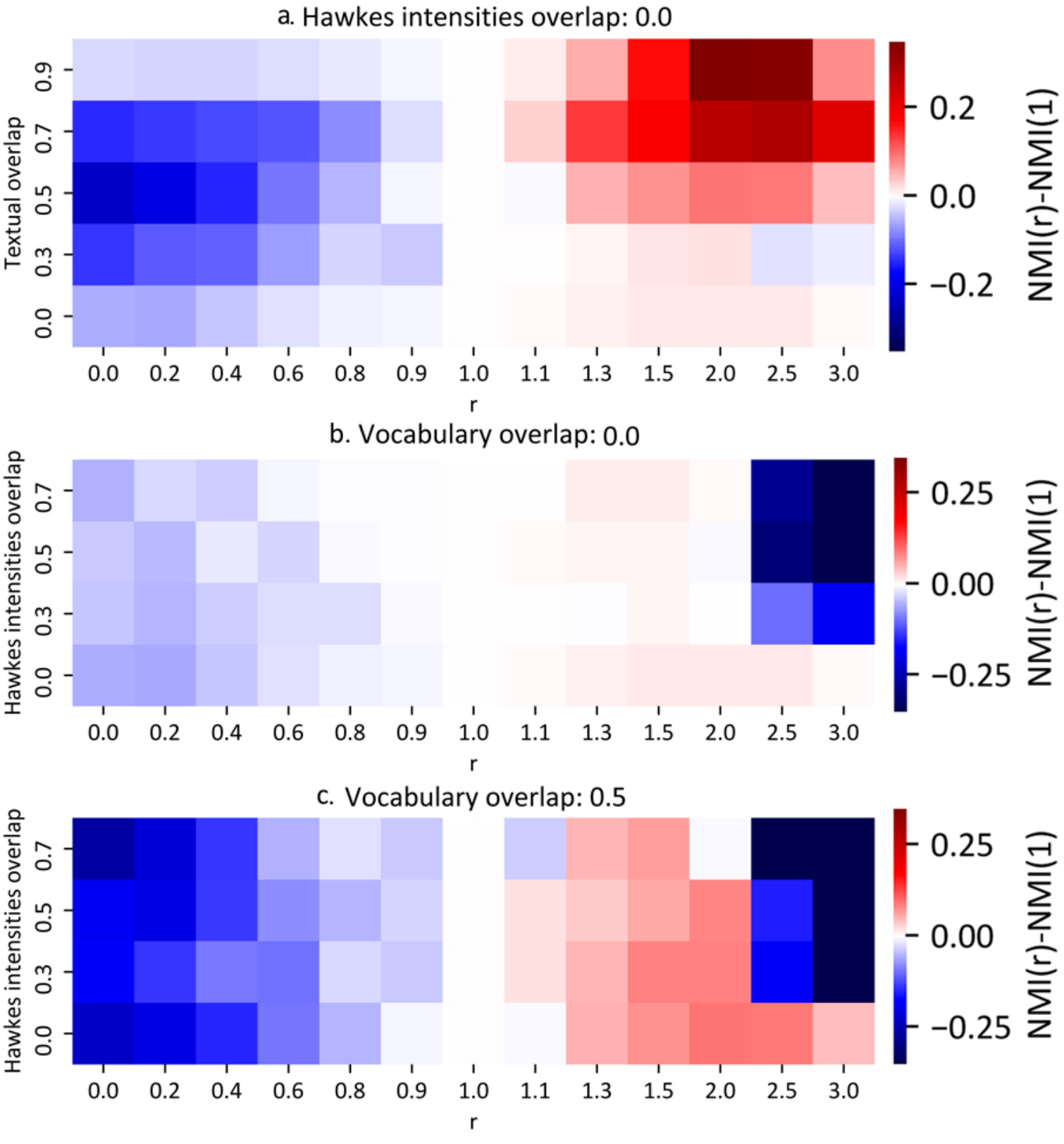}
    \caption{\textbf{PDHP performs better than DHP} --- Difference between the normalized mutual information (NMI) of PDHP and DHP model \cite{Du2015DHP} for various values of $r$, intensities overlap and vocabulary overlap, averaged over all the datasets. Red means PDHP performed better, blue means PDHP performed less well. Because PDHP($r=1$)=DHP, the column $r=1$ show no difference. PDHP allows to increase results on NMI by as much as 0.3 over DHP.}
    \label{fig-res-overlaps}
\end{figure}

\subsection{PDHP yields better results as vocabulary overlap increases}
We report our results when the intensities overlap is null, with varying $r$ and the vocabulary overlap in Fig.~\ref{fig-res-overlaps}a. Because we consider ten different datasets for each set of overlap parameters, it makes no sense to report the absolute average NMI since it can vary greatly from one dataset to the other. Instead, we plot the relative NMI difference between PDHP and DHP ($r=1$), which we expect to be less dependent on the datasets we consider. However, to give an idea of the typical performance for some parameters, we also provide raw results for one run in Fig.~\ref{fig-res-NMI}.

There is a clear correlation between efficiency, vocabulary overlap and $r$, with a gain on NMI up to $+30\%$ of its maximal value over DHP. As stated at the end of the "Model" section, this result was expected: the more vocabulary overlap grows, the less textual content carries valuable information for clustering the documents. This observation supports the concerns raised in \cite{Yin2018ShortTextDHP} about the efficiency of DHP for clustering short text documents. However, Hawkes intensities overlap being null, the arrival time of events carries highly valuable information when textual content does not allow to distinguish clusters well. Therefore, PDHP provides a way to tackle the problem raised in \cite{Yin2018ShortTextDHP} without the need to sample observations.

Conversely, when vocabulary overlap is null, the textual content provides enough information to distinguish clusters correctly. The temporal dimension only allows refining the results with no significant improvement for all values of $r$.

Finally, we can see how the Dirichlet-Uniform process (DUP, $r=0$) consistently yields worse performances under these settings. Once again, this is expected since, in this synthetic experiment, intensities overlap carry valuable information about events clustering; DUP only considers textual information and therefore misses valuable clues.

\subsection{PDHP yields similar results for null vocabulary overlap}
We report similar results in Fig.~\ref{fig-res-overlaps}b. Here, we consider a null vocabulary overlap for various values of $r$ and of Hawkes intensities overlap. The situation is now the opposite: the textual content always carries valuable information about clusters, whereas the temporal aspect does not. We observe the same trend as in Fig.~\ref{fig-res-overlaps}a --note that the color scale is the same. Varying the value of $r$ does not significantly change the performances of clustering, meaning the textual content always carries enough information. This plot shows that PDHP can handle greater intensities overlap without collapsing into unrealistic clustering. Since in most real-case applications, many clusters with various dynamics may coexist simultaneously, it is comforting that the PDHP can also handle this case.

\subsection{PDHP yields better results in more realistic situations}
We finally report the results for intermediate values of intensities and vocabulary overlaps in Fig.~\ref{fig-res-overlaps}c. In real-world applications, it seldom happens that topics vocabularies do not overlap at all. For example, a quick analysis of \textit{The Gutenberg Webster's Unabridged Dictionary by Project Gutenberg} shows that there are 22\% of English words that are associated with more than one definition. A more detailed analysis would need to consider the usage frequency of words to get correct statistics. Still, this number provides an estimate of the effective vocabulary overlap in real-world situations.

In Fig.~\ref{fig-res-overlaps}c, we present the results for a fixed vocabulary overlap of 0.5 for various values of $r$ and intensities overlap. Once again, we see that, on average, using PDHP can increase the NMI over DHP up to +20\% of the maximum possible value.

\begin{figure}
    \centering
    \includegraphics[width=\columnwidth]{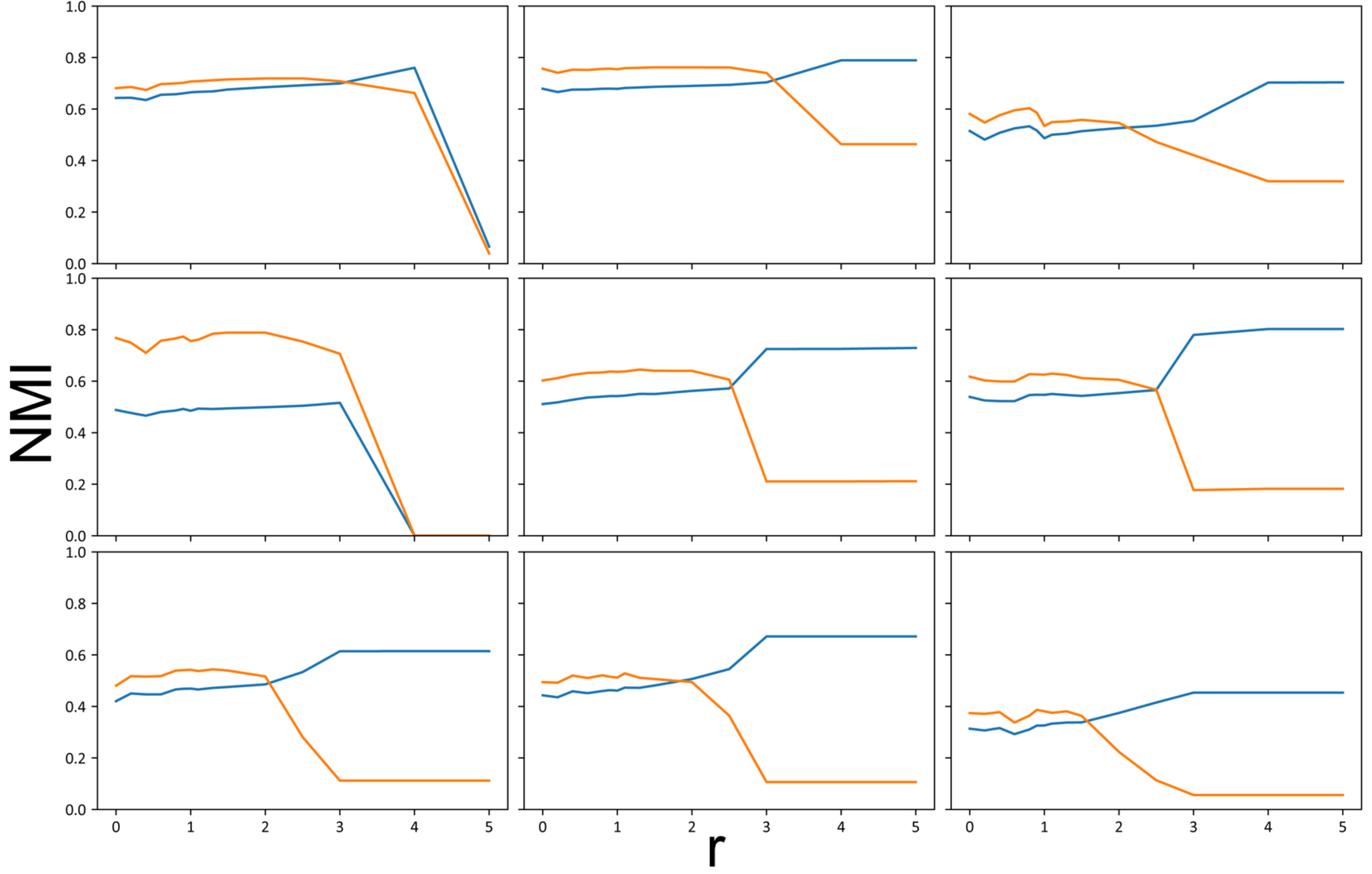}
    \caption{\textbf{Textual (orange) and temporal (blue) NMI vs r when textual and temporal clusters are decorrelated} --- From top-left to bottom-right, there are 10\%, 20\%, 30\%, 40\%, 50\%, 60\%, 70\%, 80\% and 90\% of generated events that have been randomly re-assigned a textual cluster. The orange curves are the textual NMI vs $r$, that evaluate how well events whose vocabulary has been sampled from the same distribution are clustered together; the blue curves are the temporal NMI vs $r$, that evaluate how well events following the same temporal dynamic are correctly clustered together. Values presented are for one dataset. We clearly see that varying $r$ allows to retrieve the right temporal ($r$ large) or textual clusters ($r$ small).}
    \label{fig-res-perc_rand1run}
\end{figure}
\begin{figure}
    \centering
    \includegraphics[width=\columnwidth]{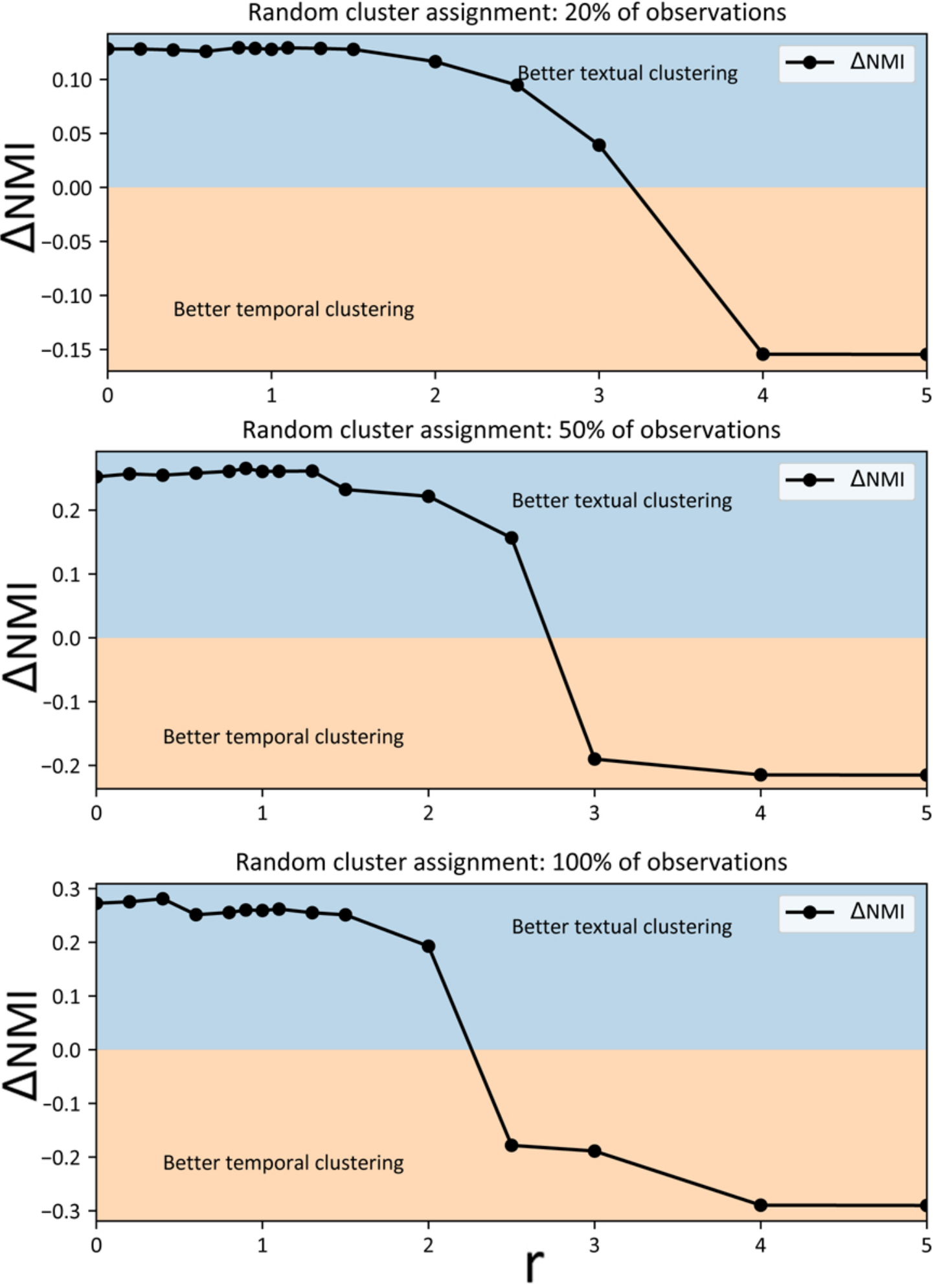}
    \caption{\textbf{Varying $r$ allows to choose between textual or temporal clustering} --- The black line plots the difference between the NMI of textual and temporal clustering. For small $r$, textual clustering is far better than temporal clustering, and for large $r$, the situation is reversed. This is because $r$ determines the importance given to the temporal dimension and therefore allows choosing between retrieving temporal or textual clusters.}
    \label{fig-res-decorr}
\end{figure}
\subsection{PDHP finds textual or temporal clusters depending on $r$}
We now slightly modify our experimental setup. Instead of considering that textual clusters and Hawkes intensities are perfectly correlated, we consider a decorrelated case. A document whose vocabulary is drawn from cluster $C_1$ can now follow the same temporal dynamics as cluster $C_2$. If we imagine a dataset of news articles published online, it is clear why this might happen frequently. If popular newspapers such as New York Times or Reuters publish an article on topic $A$ at time t, it is likely to trigger snowball publications of similar articles from less popular journals. ``Popularity'' is chosen as an indicator in this example, but it may be any other external parameter (centrality in news networks, support of publications, etc.). In this case, the article's textual content allows to uncover a ``story of publication'', that is, how the article has been spread, when publication spikes are, etc. However, the temporal information would help understand the dynamics of publications interaction: which reduced set of articles triggered the publication of subsequent ones. 

In \cite{Du2015DHP}, it is assumed that every document within clusters follow a unique dynamics. We relax this hypothesis in our datasets as follows. For null textual and temporal overlaps, after a dataset has been generated, we resample the textual clusters of a fraction of randomly selected events, as well as the words associated with the event. Doing so, we decorrelate temporal and textual clusters. Therefore, an event is now described by two cluster indicators: its temporal cluster (which Hawkes intensity made the event appear where it is) and a textual cluster (which vocabulary has been used to sample the words the event contains).

For completeness, we also show the results for for various decorrelations for one run in Fig.~\ref{fig-res-perc_rand1run}. To better understand the tendency of NMIs with respect to $r$, we plot the average difference between the NMI of textual clustering and the NMI temporal clustering over all the datasets. Explicitly: $\Delta NMI = NMI_{text}-NMI_{temp}$. The results are reported in Fig.~\ref{fig-res-decorr}.

As supposed at the end of the ``Model'' section, varying $r$ allows retrieving one clustering or the other. Note that the value $r$ of transition from text to time clustering depends directly on the dataset considered: number of words sampled, vocabulary size, overlaps, etc.

\subsection{PDHP efficiently infers the temporal dynamics of each cluster}
Finally, we mention that PDHP correctly infers kernels' parameters in every situation where events are correctly assigned to their temporal cluster. We looked at the mean absolute error (MAE) between the vector $\vec{\alpha}$ used to generate the dataset and the inferred one. When documents are correctly classified, the MAE according to the actual $\vec{\alpha}$ entries is systematically lower than 0.1. We do not discuss this metric further because it is directly correlated to the NMI metric. If documents are correctly classified, the inferred intensity function is based on the correct observations and corresponds to the one used for data generation. If documents are not correctly classified, the inferred intensity function is close to the optimal one ($\sim 5\%$ MAE) given the available information but may be far from the one used to generate datasets because events are misclassified.

\subsection{Real-world application on Reddit}
\begin{figure}
    \centering
    \includegraphics[width=\columnwidth]{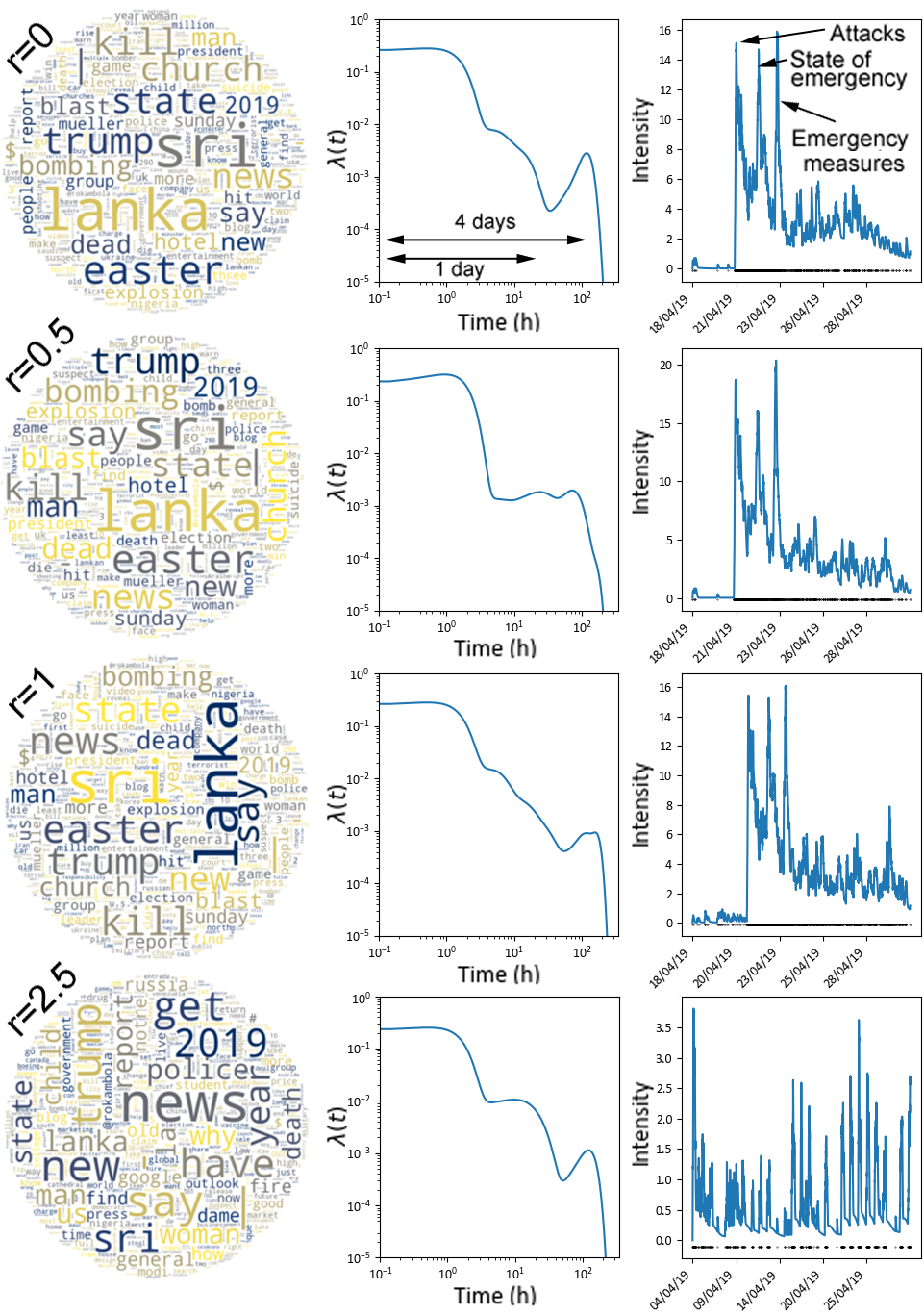}
    \caption{Wordclouds, triggering kernels and intensities for clusters the most closely related to Sri Lanka 2019 bombings for various values of $r$. The points at the bottom of the intensity plots are individual publication events. Note that triggering kernels are plot on a log-log scale for visualisation purpose, because most of the intensity is focused on small times: dynamics are bursty.}
    \label{fig-res-Srilanka}
\end{figure}
We use the PDHP prior to model real streams of textual documents. We consider three Reddit datasets\footnote{Available for download at https://files.pushshift.io/reddit/submissions/} about different topics. The \textbf{News dataset} is made of 73.000 titles extracted from the subreddits inthenews, neutralnews, news, nottheonion, offbeat, open\_news, qualitynews, truenews and worldnews, from April 2019. We chose this month because of the wide variety of events that happened then (for instance, Sri Lanka Easter bombings, Julian Assange arrest, first direct picture of a black hole, Notre-Dame cathedral fire). We also consider 15.000 post titles of the subreddit TodayILearned (\textbf{TIL dataset}) and 13.000 post titles of the subreddit AskScience (\textbf{AskScience dataset}) on January 2019. We extracted the nouns, verbs, adjectives and symbols from the textual data. We run the experiments using the following parameters: $\alpha_0=0.5$, $\theta_0=0.01$, $N_{samples}=2.000$, $N_{part}=8$ and $\omega_{thres}=\frac{1}{2N_{part}}$. The kernel vector $\vec{\kappa}$ is made of Gaussian functions, with means located at 0.5, 1, 4, 8, 12, 24, 48, 72, 96, 120, 144 and 168 hours. The variance of each are set to 1, 1, 3, 8, 12, 12, 24, 24, 24, 24, 24 and 24 hours. The algorithm will then infer the weights $\vec{\alpha}$ associated with each entry of the kernel vector $\vec{\kappa}$ for each cluster.

\begin{figure}
    \centering
    \includegraphics[width=\columnwidth]{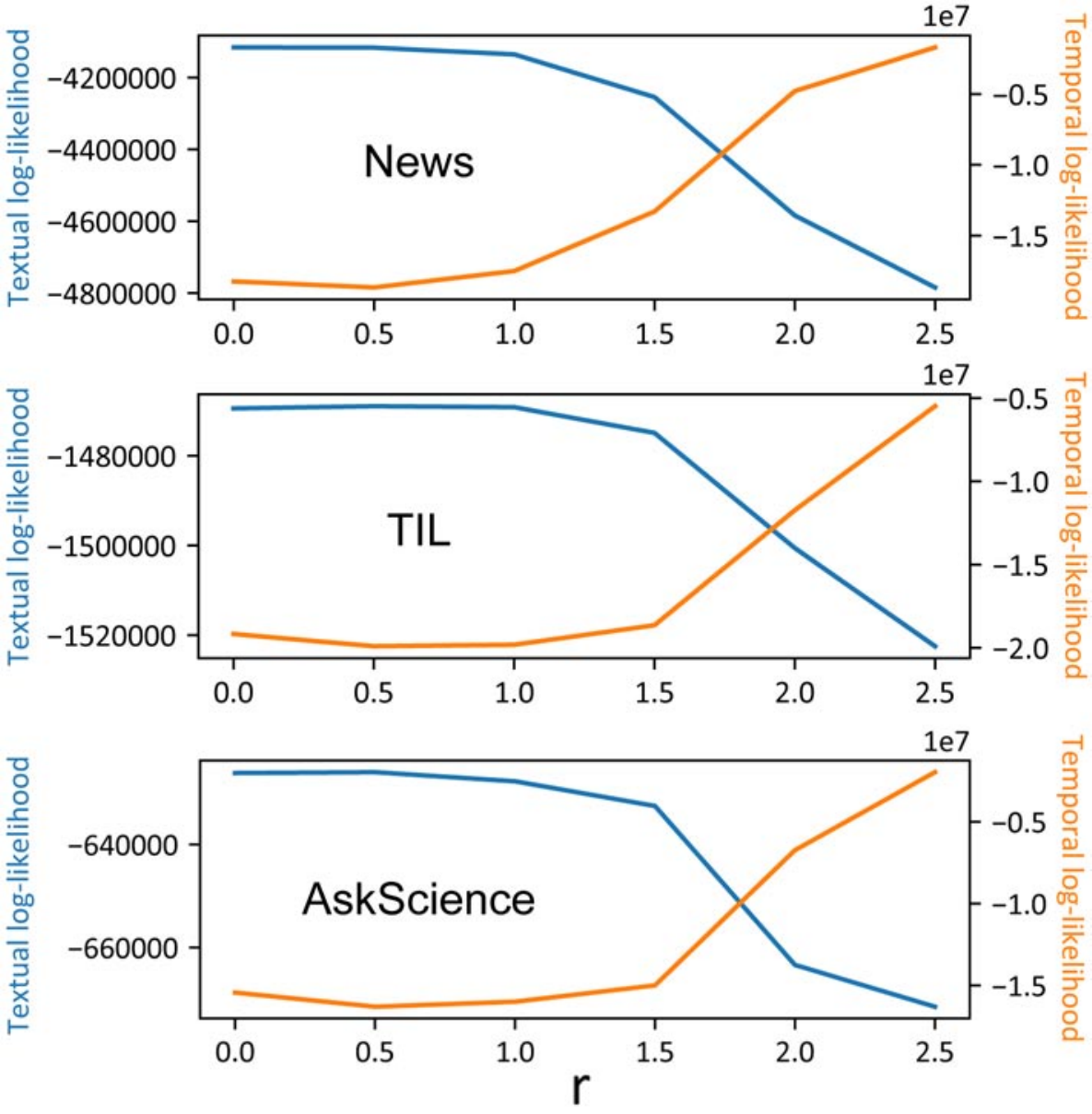}
    \caption{\textbf{$r$ allows to favour text-based of time-based clustering on real world datasets} --- Textual likelihood and Hawkes process likelihood for various values of $r$. The lower $r$ the higher textual likelihood is, and the higher $r$ the higher Hawkes process likelihood is.}
    \label{fig-likReddit}
\end{figure}
\subsubsection{PDHP recovers meaningful stories}
As an illustrative example, we consider the inferred clusters the most related to Sri Lanka easter bombings of April 21$^{st}$ 2019 in Fig.\ref{fig-res-Srilanka}. We plot the triggering kernels on a log-log scale, because most of the intensity is focused on small times: dynamics of information spread are bursty \cite{Karsai2018BurstyHumanDynamics}. We see that inferred dynamics change with $r$, which is expected since clusters do not contain the same documents. For $r=0$, the Uniform process makes clusters based on textual information only; the triggering kernel is inferred afterward. For $r=2.5$ on the contrary, clusters are formed based on the triggering kernel, and textual information follows; we see from the right-plot that this cluster captures publications exhibiting a daily intensity cycle. Given the intensity spikes on 21$^{st}$, 22$^{nd}$, and 23$^{rd}$, it is not surprising that articles about Sri Lanka bombings are also part of this cluster. Note that the more $r$ increases, the more intense the triggering kernel is around 24h. We see from Fig.\ref{fig-res-Srilanka} that DHP is a specific case of the dynamics one can retrieve using temporal information.

\subsubsection{PDHP favors temporal or textual clustering depending on $r$}
We report the values of log-likelihoods for every dataset and various values of $r$ in Fig.\ref{fig-likReddit}. The textual likelihood is defined Eq.\ref{eq-likModelLg}, and the likelihood of a Hawkes process is defined Eq.\ref{eq-likHawkes}. Note that $r$ does not appear in any of these expressions. Those likelihoods evaluate how well the textual or temporal aspect of the dataset is modeled without considering the PDHP process. As expected, varying $r$ makes the model more sensitive to either textual or temporal data ---low $r$ favors a text-based clustering, whereas high $r$ favors a time-based clustering.

\begin{figure}
    \centering
    \includegraphics[width=0.8\columnwidth]{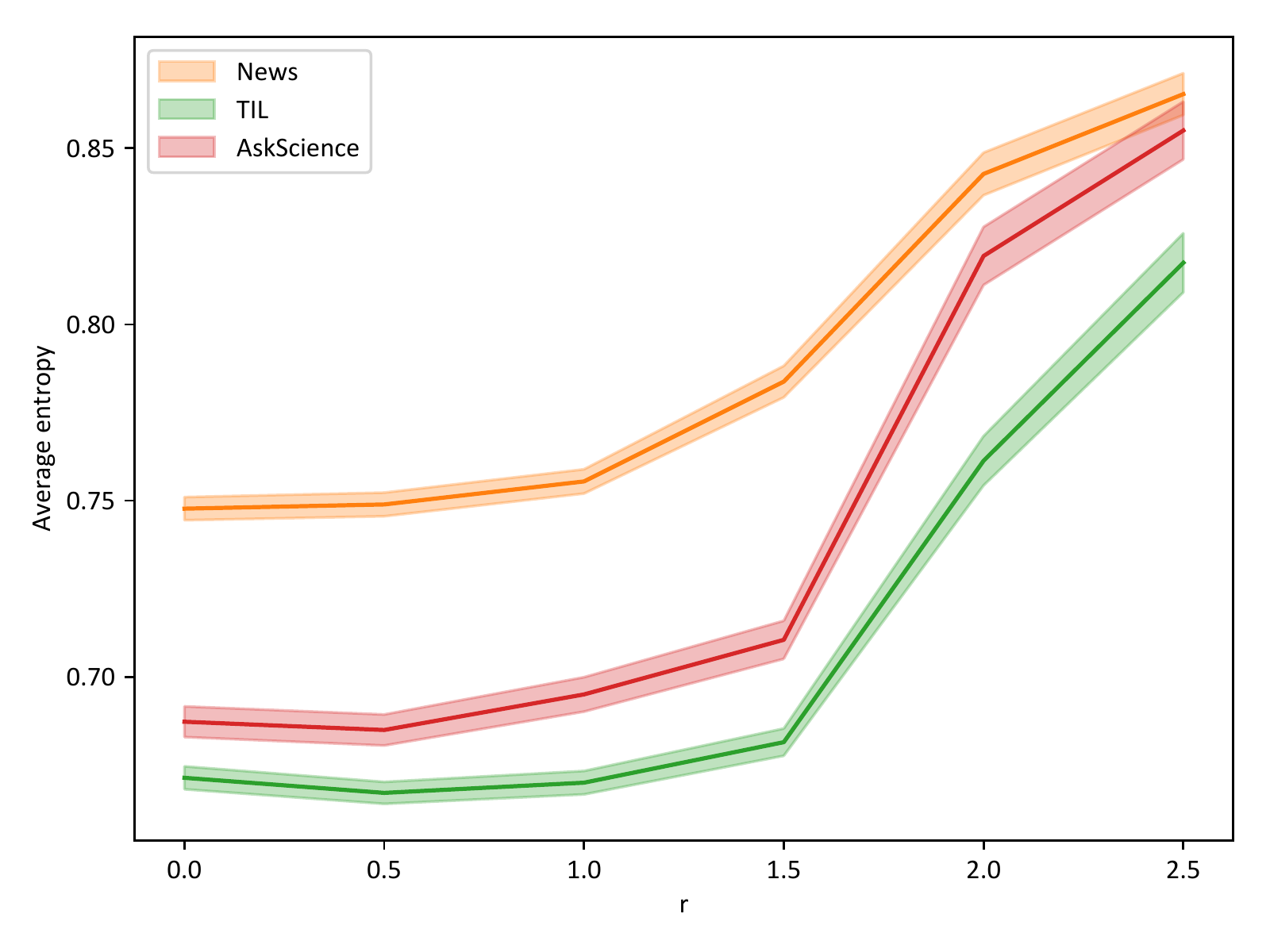}
    \caption{\textbf{Textual clusters are more informative for low values of $r$} --- Weighted average entropy of words distribution for every dataset. Weights corresponds to the number of words within clusters. The error bar represents the standard error over all the clusters.}
    \label{fig-entropy-txt}
\end{figure}
\subsubsection{PDHP infers sharper textual clusters for low $r$}
We evaluate how meaningful textual cluster are using entropy. We would like a set of words within a cluster to carry a meaningful information. A way to measure this is to see how spread the vocabulary of a cluster is. Let $N_{c,v}$ be the count of word $v$ in cluster $c$. The normalized Shannon entropy of a cluster $c$ is defined as:
\begin{equation}
    \label{eq-Entropy}
    S(\vec{N_c}) = \frac{1}{-\log (V)}\sum_v^V \log (\frac{N_{c,v}}{\sum_v' N_{c,v'}})\frac{N_{c,v}}{\sum_v' N_{c,v'}}
\end{equation}
An entropy of 0 means the vocabulary of the cluster is concentrated on a single word; an entropy of 1 means that every of the $V$ words is present to the same extent. In Fig.\ref{fig-entropy-txt}, we plot the mean entropy for various values of $r$ for all the datasets, along with the standard error over the clusters. The results show that vocabulary is more concentrated within clusters for low values of $r$. The inflection point of the curves corresponds to what has been previously observed with likelihoods in Fig.\ref{fig-likReddit}. On the contrary, higher values of $r$ lead to clusters that comprise a less dense vocabulary. This is expected because as $r$ increases, the textual information is no longer the most relevant data for cluster formation.

\section{Conclusion}
We built the Powered Dirichlet-Hawkes process as a generalization of the Dirichlet-Hawkes process and Uniform process and showed how it improves performance on various datasets. When textual information conveys little information, or when temporal information conveys little information, and when both do, our model is able to correctly retrieve the original clusters used in the generation process with high accuracy.
A central consideration in document clustering is that there are no ``right'' clusters. For instance, we illustrate how textual content and temporal dynamics can be decorrelated in real-life applications. The framework we developed is flexible enough to allow users to choose the weight they wish to give to temporal or textual information depending on the situation; when textual and temporal clusters are decorrelated, the model allows one to choose which of those to infer.

Many future extensions are possible for PDHP. For instance, it would be interesting to develop its hierarchical version (PHDHP) as it has already been done with HDHP for DHP. Another interesting perspective would be to create a version considering multivariate Hawkes processes to study how textual clusters' dynamics relate to each other. Given several recent works have been based on the regular Dirichlet-Hawkes process, it would be insightful to study how their results vary when using the Powered Hawkes-Dirichlet process instead. A study on the influence of the language model used along with PDHP would also be interesting since the text model we used here was simple on purpose (our focus being on the PDHP prior and not on the model it gets associated with).

Finally, it would be interesting to see how this model would work in another context where temporal and textual information are intertwined. For instance, in latent social network inference, we may be able to create clusters according to the observed temporal dynamics of publications, or according to the textual information shared between users, or according to a combination of both.

\bibliographystyle{IEEEtran}
\bibliography{Bibliography.bib}

\end{document}